\documentclass[letterpaper, 10 pt, conference]{ieeeconf}  

\IEEEoverridecommandlockouts                              

\usepackage{graphics} 
\usepackage{epstopdf}
\usepackage{epsfig} 
\usepackage{subfigure}
\usepackage{algorithm}
\usepackage{algorithmicx}
\usepackage{algpseudocode}
\usepackage{multirow}
\usepackage{booktabs}
\usepackage{amsmath} 
\usepackage{amssymb}  

\title{\LARGE \bf
Towards More Realistic Human-Robot Conversation: A Seq2Seq-based Body Gesture Interaction System
}

\author{Minjie Hua$^{1}$, Fuyuan Shi$^{1}$, Yibing Nan$^{1}$, Kai Wang$^{1}$, Hao Chen$^{1}$, and Shiguo Lian$^{1}$
\thanks{$^{1}$ All the authors are with CloudMinds Technologies Inc., Beijing 100102, China.
        {\tt\small michael.hua, fuyuan.shi,
        charlie.nan, kai.wang, hao.chen,
        scott.lian@cloudminds.com}}%
}

\begin{document}

\maketitle
\thispagestyle{empty}
\pagestyle{empty}

\begin{abstract}
This paper presents a novel system that enables intelligent robots to exhibit realistic body gestures while communicating with humans. The proposed system consists of a listening model and a speaking model used in corresponding conversational phases. Both models are adapted from the sequence-to-sequence (seq2seq) architecture to synthesize body gestures represented by the movements of twelve upper-body keypoints. All the extracted 2D keypoints are firstly 3D-transformed, then rotated and normalized to discard irrelevant information. Substantial videos of human conversations from Youtube are collected and preprocessed to train the listening and speaking models separately, after which the two models are evaluated using metrics of mean squared error (MSE) and cosine similarity on the test dataset. The tuned system is implemented to drive a virtual avatar as well as Pepper, a physical humanoid robot, to demonstrate the improvement on conversational interaction abilities of our method in practice.
\end{abstract}

\section{Introduction}\label{sec:intro}

A conversation system is one of the most essential modules for intelligent robots. Current intelligent robots are already capable of receiving prosodic signals from humans and giving appropriate verbal responses with techniques proposed in~\cite{HG12,GM13,SV14,FQ14,ZS15}, but few of them can express nonverbal feedback while listening, or present variable body gestures according to the verbal responses while speaking, which makes the human-robot communication seemed unnatural.

The authors in~\cite{MJ12} enable the robots to exhibit body gestures during conversations. However, the possible options are pre-defined and limited, which means that the robot motions are constrained. Moreover, as far as we know, it lacks of work aiming at synthesizing body gestures for robots during the listening phase in their talks with human.

\begin{figure}[tbp]%
\centering
\includegraphics[scale=0.5]{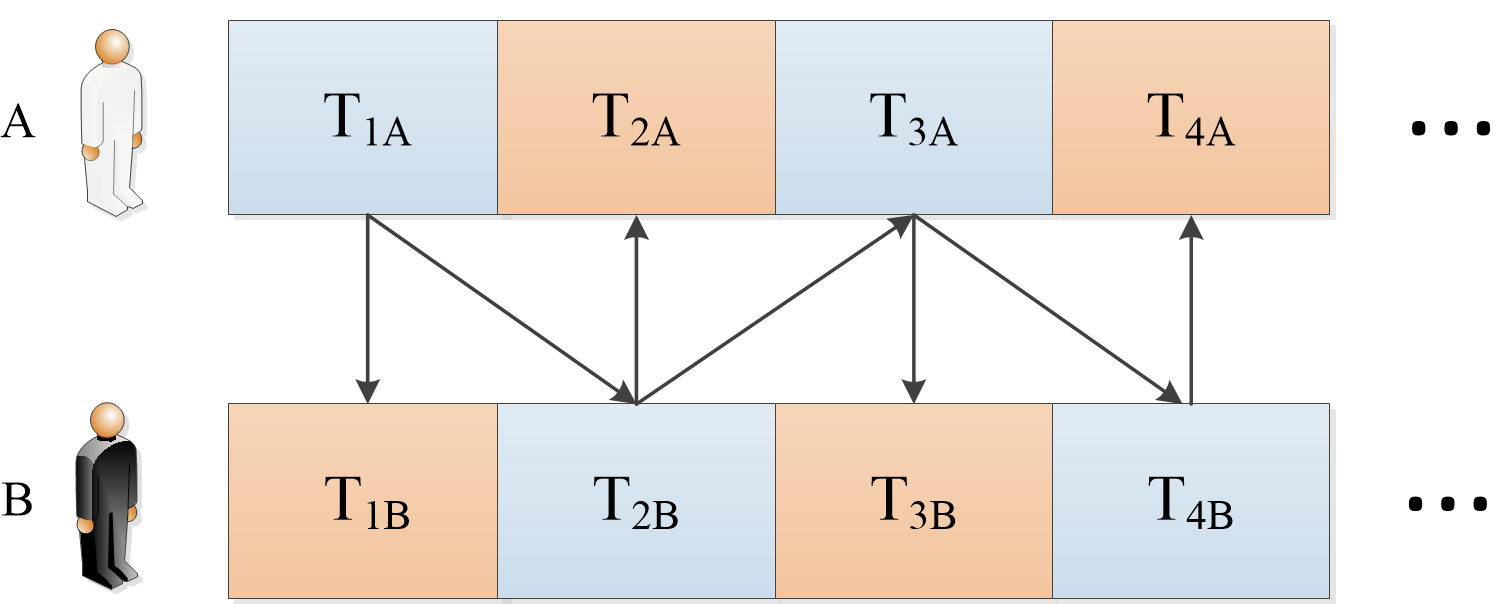}
\caption{Regularity of human-human conversation. $\text{T}_{i\text{A}}$ and $\text{T}_{i\text{B}}$ ($i$ = 1, 2, 3, ...) denotes the reactions of person A and B respectively in the $i$-th time period. Blue blocks indicate that the person is speaking with body gestures. Orange ones indicate that the person is listening and may give nonverbal feedback. Arrows imply the dependence relationship.}
\label{fig:regularity}
\end{figure}

Our research is based on the observations of conversational regularity illustrated in Fig.~\ref{fig:regularity}: In human-human communications, the two roles, listener and speaker, alternate between the two parties involved in the conversation. While the listener is listening, he/she receives utterance signals as well as nonverbal cues from the speaker and may give nonverbal feedback in the meantime. Their roles exchange when the previous speaker stops talking. Then the new speaker makes verbal responses with appropriate body gestures based on what he/she heard. This procedure carries on repeatedly until the conversation ends.

In this paper, we aim to enhance the abilities of intelligent agents such as virtual avatar and real-world humanoid robots with better comprehension and expression of body gestures. It will not only help them appear more expressive, intelligible and interactive, but also provide humans with a more natural communication experience with robots.

Inspired by the great success of sequence-to-sequence (seq2seq) network~\cite{SV14} in the sequence mapping problems, we propose a human-robot interaction system composed of seq2seq-based listening and speaking models for synthesizing body gestures. The listening model takes both the speaker's verbal and nonverbal signals as input and generates body gestures as nonverbal feedback. And the speaking model takes only the verbal response as input and generates body gestures as nonverbal accompaniment.

The prerequisite of our system is to transform utterance and body gesture into numerical features. Speech recognition algorithms~\cite{HG12,GM13} convert utterance to text, which can be encoded to a vector sequence using word embedding algorithms~\cite{MS13,MC13,PS14,PM18,DC18}. As for body gesture, the models proposed in~\cite{OP18,FX17,TR17} extract 2D coordinates of body keypoints and transform them into 3D space. However, the coordinate representation includes much noise, so we develop keypoints rotation and normalization methods in the gesture parsing module to discard irrelevant information. To demonstrate the prediction of our system, the body gestures generated by the listening and speaking models are reconstructed on avatar or robot by motion synthesis module.

The rest of the paper is organized as follows: Section~\ref{sec:review} reviews related work on conversational systems and nonverbal expressions synthesis models. The architecture of the proposed system is presented in Section~\ref{sec:sys}. In Section~\ref{sec:key}, the body gesture parsing process is described in detail. Section~\ref{sec:model} introduces the seq2seq-based listening model and speaking model that realize body gestures generation. Experimental results are given and discussed in Section~\ref{sec:exp}. And finally, Section~\ref{sec:conc} gives the conclusion.

\begin{figure*}[htbp]%
\centering
\includegraphics[scale=0.6]{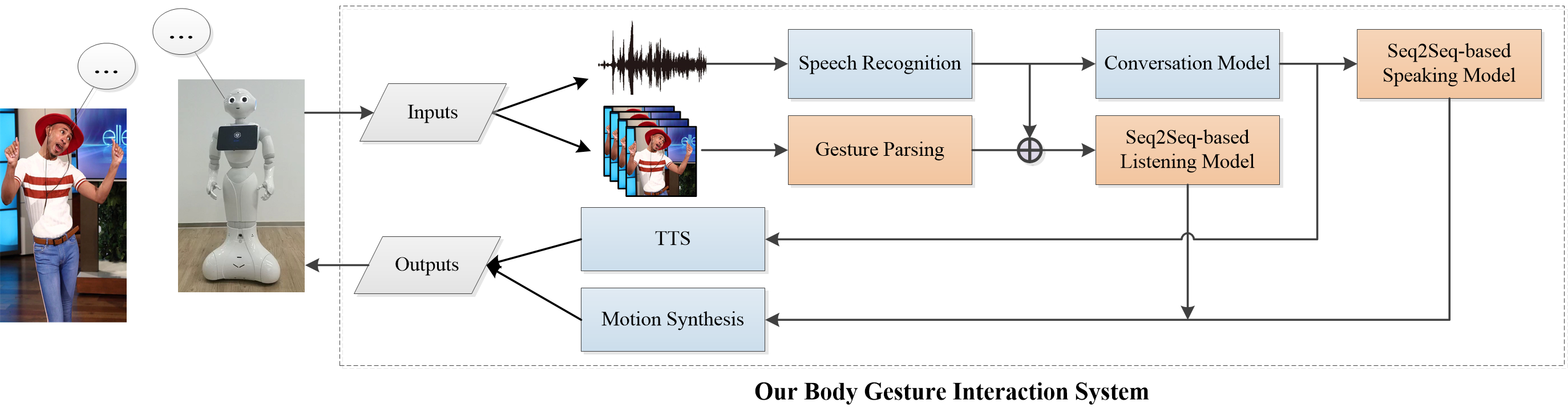}
\caption{Overview of the proposed human-robot body gesture interaction system. Blocks colored orange indicate contributions of our work.}
\label{fig:overview}
\end{figure*}

\section{Related Work}\label{sec:review}

Conversational systems have been explored for many years. Early dialogue models like ELIZA~\cite{WJ66} and PARRY~\cite{CK81} can already respond to relatively complicated questions based mainly on hand-crafted rules, which makes the conversation seemed somewhat monotonous. The authors in~\cite{RC11} pioneered to formulate this problem as language translation. However, they met some difficulties because the space of possible responses in conversation is much larger than that in language translation.

The authors in~\cite{SV14} developed seq2seq framework, which was applied to machine translation and achieved excellent performance. Later, they introduced the same approach to conversational modeling in~\cite{VL15}. In analogy to mapping a sentence from one language to another in machine translation, the conversational model maps a query sentence to a response sentence. Generally, seq2seq framework uses an LSTM~\cite{HS97} layer to encode the input sentence to a vector of fixed dimensionality, and then another LSTM layer to decode the target sentence from the vector. This encoder-decoder architecture is widely used in sequence mapping problems like machine translation~\cite{SV14}, conversation modeling~\cite{VL15} and even video description~\cite{VR15} due to its powerful capabilities.

Aiming at producing more diverse responses in original seq2seq-based conversation model, \cite{LG15} modified the object function to encourage variety. The authors in~\cite{SS16} extended hierarchical recurrent encoder-decoder neural network to conversational modeling and upgraded LSTM units to advanced GRU units~\cite{CG14}. They improved the model further in~\cite{SI17} by appending stochastic latent variables to generate more diverse and meaningful responses.

All the aforementioned methods involve only verbal information. However, nonverbal expressions like body gestures and facial movements are commonly used in human conversations. In recent years, facial gestures synthesis domain gains much attention. \cite{HL17} presented a method that can reconstruct appearance-like virtual heads from a single RGB image of humans. \cite{SS15} captured physical features from a huge collection of photos of a person, and reconstructed Tom Hanks, an avatar, from the learned personal characteristics. \cite{TZ16} proposed an approach for making the person in target video reenact the facial expressions of another person captured with a webcam. \cite{SS17} synthesized realistic facial expressions and lip sync for a talking avatar from audio signals using an RNN network. Furthermore, the authors in~\cite{CL18} fused facial cues into conversation model based on the observation that the same sentence might have different meanings with different sentiments conveyed by facial gestures. They adopted RNN encoder-decoder architecture to generate both verbal responses and facial expressions for a chatting avatar.

In comparison with facial gestures, research on body gesture synthesis is left behind. \cite{KI03} proposed an exploratory analysis of body gesture interaction between humanoid robots and humans and proved that arm movements play an important role in conversations. Authors in~\cite{LT09, LK10} exploited a system that synthesizes appropriate body gestures based on prosodic features extracted from real-world speech using hidden Markov model (HMM). However, they aimed at enhancing the avatar's performance in virtual environment, thus concentrated little on human-robot conversational interactions. \cite{MJ12} presented an approach to extend communicative behavior for Nao, a humanoid robot, with a set of pre-defined body gestures. In~\cite{AT12, AT16}, the authors utilized the coupled hidden Markov models (CHMM) to generate verbal responses accompanied with arm movements based on the human's prosodic characteristics. In this paper, a body gesture interaction system is constructed based on seq2seq network to provide more natural human-robot conversational experiences. In this system, the body gestures of avatar or robot are driven by the models trained with the video data captured from real human-human conversations.

\section{System Overview}\label{sec:sys}

\begin{figure*}[htbp]%
\centering
\includegraphics[scale=0.35]{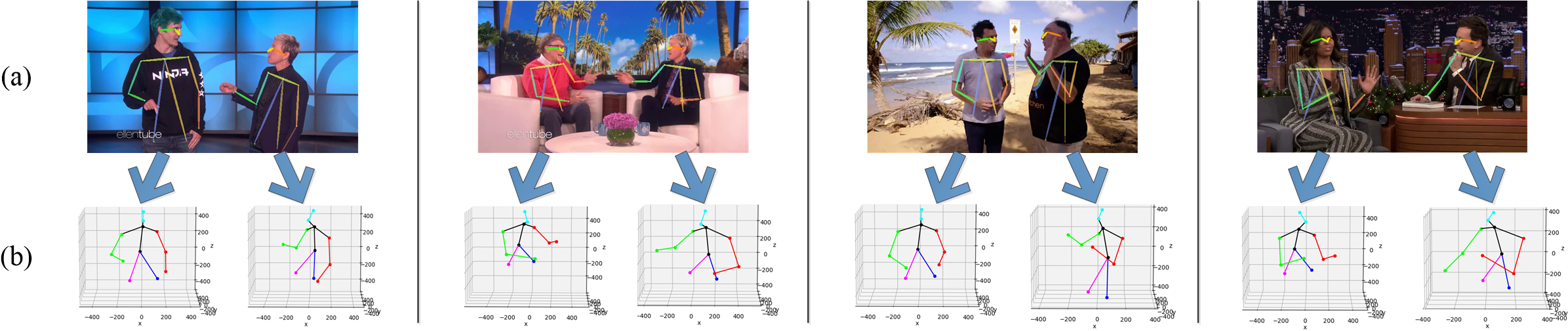}
\caption{Examples of keypoints extraction and transformation from 2D to 3D. In row (a), adjacent connections of 2D keypoints are drawn on detected persons. And the scatterplots in row (b) are the transformed 3D keypoints skeleton.}
\label{fig:keypoints}
\end{figure*}

In this paper, we propose a novel human-robot interaction system with seq2seq-based body gesture generation models. As shown in Fig.~\ref{fig:overview}, when the robot is communicating with a human, audio signals containing verbal information and RGB frames containing nonverbal information are input to our system. The audio is then transformed to text using the speech recognition algorithm proposed in~\cite{HG12, GM13} and the raw frames are processed by the body gesture parsing module, which will be elaborated in the next section. The extracted text is branched into two routes: one is passed to the conversation model for generating response sentence, and another, along with the parsed body gesture, is input to the listening model. Then the listening model predicts output body keypoints sequence, which is analyzed by the motion synthesis model to perform body gesture feedback while the robot is listening. When the human stops talking, the response sentence is transformed to prosodic signals using text-to-speech (TTS) model proposed in~\cite{FQ14, ZS15}. In the meantime, the speaking model generates body keypoints sequence based on the response sentence. With the motion synthesis module, the robot is capable of giving both verbal and nonverbal responses.

\section{Gesture Parsing}\label{sec:key}

\subsection{Dataset Overview}

Our listening and speaking models needed substantial data of human-human conversation to synthesize realistic body gestures. In addition, considering that the behavioral habits differ among humans, we hoped that one of the communicators is fixed during the training process to ensure the persona consistence.

We found talk shows like The Ellen Show and The Tonight Show meet our requirements perfectly. We downloaded 2263 videos of The Ellen Show and 1978 videos of The Tonight Show from Youtube. All the collected videos were segmented into clips based on the audio signals. Using the speaker recognition method proposed in~\cite{DG85}, we could distinguish the current speaker and cut the video at each role-exchange border. Then, the audio in each clip was transferred to corresponding text by speech recognition module. As a result, we got several clip-text pairs with the ID of recognized speaker included in the text. Finally, all the clips were processed by gesture parsing module, which is composed of the keypoints extraction, 3D-transformation, rotation and normalization.

\subsection{Keypoints Extraction}

Benefitting from the outstanding performance of AlphaPose~\cite{FX17}, the module could easily extract each person's 2D keypoints from each frame. At first, we attempted to focus on 17 keypoints of the whole body. However, we discovered that human's lower body was usually invisible in many videos. Moreover, even when the lower body appears, it always seems stable because people rarely move their positions while communicating with others. Consequently, ignoring the lower-body keypoints would largely increase the number of usable videos without causing much deviation. Fig.~\ref{fig:keypoints}(a) shows the extraction of 2D keypoints.

In the dataset preparation phase, we also discarded the clips that did not contain exactly two people. After that, we obtain 52403 clips of The Ellen Show and 51347 clips of The Tonight Show in total.

\subsection{Keypoints Transformation from 2D to 3D}

It was insufficient to reconstruct body gestures from 2D keypoints only. Using~\cite{TR17}, we obtained 3D keypoints coordinates by inputting original RGB frame together with 2D coordinates extracted by AlphaPose into the model proposed in~\cite{TR17}. Then the corresponding 3D keypoints coordinates of each person were estimated individually (see Fig.~\ref{fig:keypoints}(b)).

In our experiments, twelve 3D keypoints were selected as follows: head top, neck, chest, belly, left and right shoulders, elbows, wrists and hips. Since we focused mainly on the body gestures, facial keypoints were simplified to head top and neck for locating the position of head.

After this step, each keypoints group was represented by a $3\times 12$ matrix $\mathbf{K}=\left[\mathbf{k}_1, \mathbf{k}_2, \mathbf{k}_3, ..., \mathbf{k}_{12} \right]$, where $\mathbf{k}_i= (x_i, y_i, z_i)^T$ indicates the 3D coordinates of the $i$-th  keypoint.

\subsection{Keypoints Rotation}

We recognized that similar body gestures might differ completely in their representations because of the camera perspective. To eliminate this effect, all the keypoints groups were rotated.

We exploited our keypoints rotation algorithm based on the hypothesis that the shoulders on the both sides are at the same height. In other words, the $z$-coordinates of left and right shoulders are expected to be equal. This hypothesis was proved reasonable as we analyzed all the clips and found that in more than $90\%$ keypoints groups, the difference between their left and right shoulders' $z$-coordinates was less than $10\%$ of their heights.

\begin{figure*}[htbp]%
\centering
\includegraphics[scale=0.39]{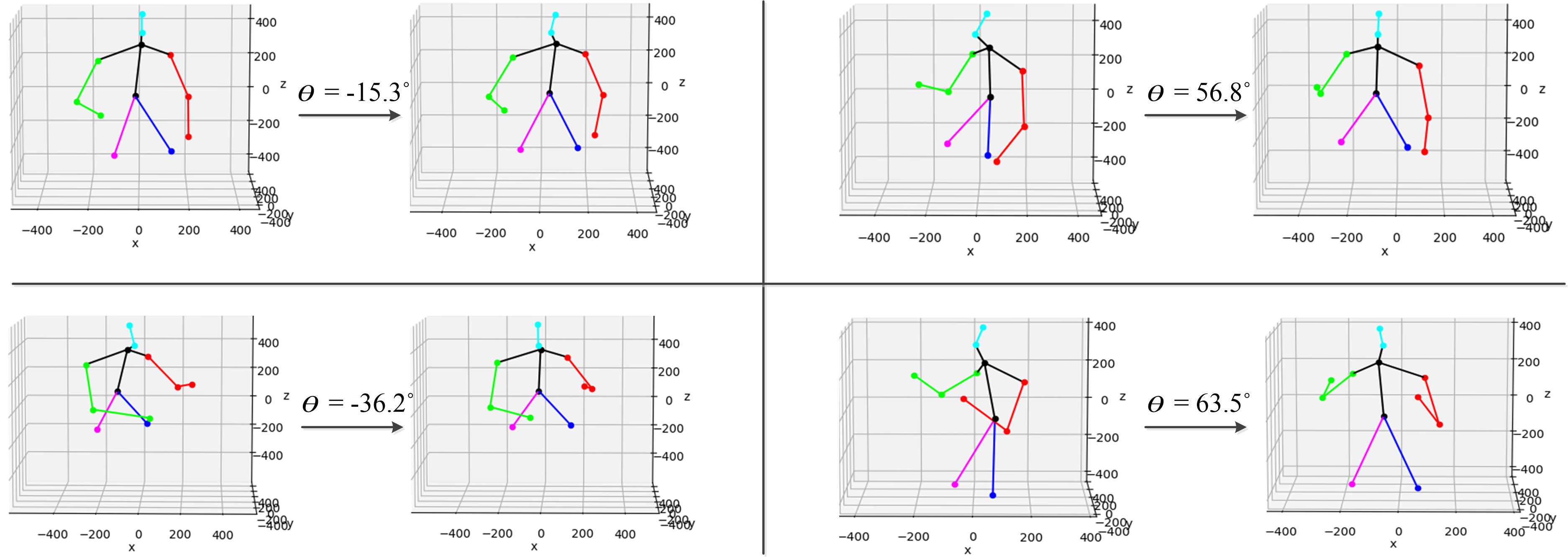}
\caption{Four examples of keypoints rotation, where $\theta$ denotes the rotated degree about $z$-axis anticlockwise}
\label{fig:keypointsrot}
\end{figure*}

Based on the above hypothesis, we rotated all the keypoints groups around $z$-axis. Suppose that the $i$-th keypoints matrix is $\mathbf{K}^i$, where the coordinates of the left and right shoulders are $\mathbf{k}^i_l= (x^i_l, y^i_l, z^i_l)$ and $\mathbf{k}^i_r= (x^i_r, y^i_r, z^i_r)$ respectively. It can be assumed that $z^i_l \approx z^i_r$ through the hypothesis. For the $i$-th keypoints group, we hoped that $y^i_l = y^i_r$ and $x^i_l > x^i_r$ are satisfied after the rotation by an angle $\theta^i$ anticlockwise (see Fig.~\ref{fig:keypointsrot}). We first obtained the difference vector $\mathbf{k}^i_{diff}=\mathbf{k}^i_l-\mathbf{k}^i_r$ and then calculated $\theta^i$ as the angle between $x$-axis and the projection of $\mathbf{k}^i_{diff}$ on $XOY$ plane. As $\theta^i$ is determined, the $i$-th rotated keypoints matrix $\overline {\mathbf{K}^i}$ is calculated by:

\begin{equation}\label{eq:k0}
\begin{aligned}
 \overline {\mathbf{K}^i}=\mathbf{R}_z(\theta^i)\mathbf{K}^i,
\end{aligned}
\end{equation}
where $\mathbf{R}_z(\alpha)$, the basic rotation matrix about $z$-axis, is defined as:

\begin{equation}\label{eq:rz}
\begin{aligned}
\mathbf{R}_z(\alpha)=
\left[ {\begin{array}{*{20}c}
   {\cos\alpha} & {-\sin\alpha} & {0}  \\
   {\sin\alpha} & {\cos\alpha} & {0}  \\
   {0} & {0} & {1}  \\
\end{array}} \right]
\end{aligned}
\end{equation}

\subsection{Keypoints Normalization}

In most machine learning algorithms, data normalization is one of the most important steps in data processing. It standardizes different scales of features, which has been proved helpful in promoting convergence of neural networks.

The goal of keypoints normalization was to eliminate the effects of absolute body positions and scales under the principle that the same body gestures should have the same representations. We attempted three normalization methods in our experiments, and the comparison among them is elaborated in Section~\ref{sec:exp}.

\subsubsection{Individual normalization}

For $i$-th rotated keypoints matrix $\overline {\mathbf{K}^i}=\left[\overline{\mathbf{k}^i_1}, \overline{\mathbf{k}^i_2}, \overline{\mathbf{k}^i_3}, ..., \overline{\mathbf{k}^i_{12}} \right]$, where $\overline{\mathbf{k}^i_j}= (\overline{x^i_j}, \overline{y^i_j}, \overline{z^i_j})^T$, we denote $x^i_{\max}$ and $x^i_{\min}$ as the maximum and minimum of $\overline{x^i_j}$ in $\overline {\mathbf{K}^i}$. $y^i_{\max}$, $y^i_{\min}$, $z^i_{\max}$ and $z^i_{\min}$ are defined similarly.

Using individual normalization, the $i$-th normalized keypoints matrix $\widetilde {\mathbf{K}^i}$ was calculated by:

\begin{equation}\label{eq:k0tilde}
\begin{aligned}
 \widetilde{\mathbf{K}^i } = \left( {\overline {{\mathbf{K}}^i }  - {\mathbf{K}}^i_{\min } } \right) \odot {\mathbf{K}}^i_{{\rm{scale}}},
\end{aligned}
\end{equation}
where ${{\mathbf{K}}^i_{\min }}$ is composed of 12 columns of ${{\left( {{x}^i_{\min }},{{y}^i_{\min }},{{z}^i_{\min }} \right)}^{T}}$, ${{\mathbf{K}}^i_{\rm{scale}}}$ is composed of 12 columns of ${{\left( ({{{x}^i_{\max }}-{{x}^i_{\min }}})^{-1},({{{y}^i_{\max }}-{{y}^i_{\min }}})^{-1},({{{z}^i_{\max }}-{{z}^i_{\min }}})^{-1} \right)}^{T}}$and $\odot$  denotes the element-wise multiplication of two matrices.

\begin{figure} [bp]
  \centering
  \subfigure[]{
    \centering
    \label{fig:indifail:a} 
    \begin{minipage}[b]{0.15\textwidth}
      \centering
      \includegraphics[scale=0.3]{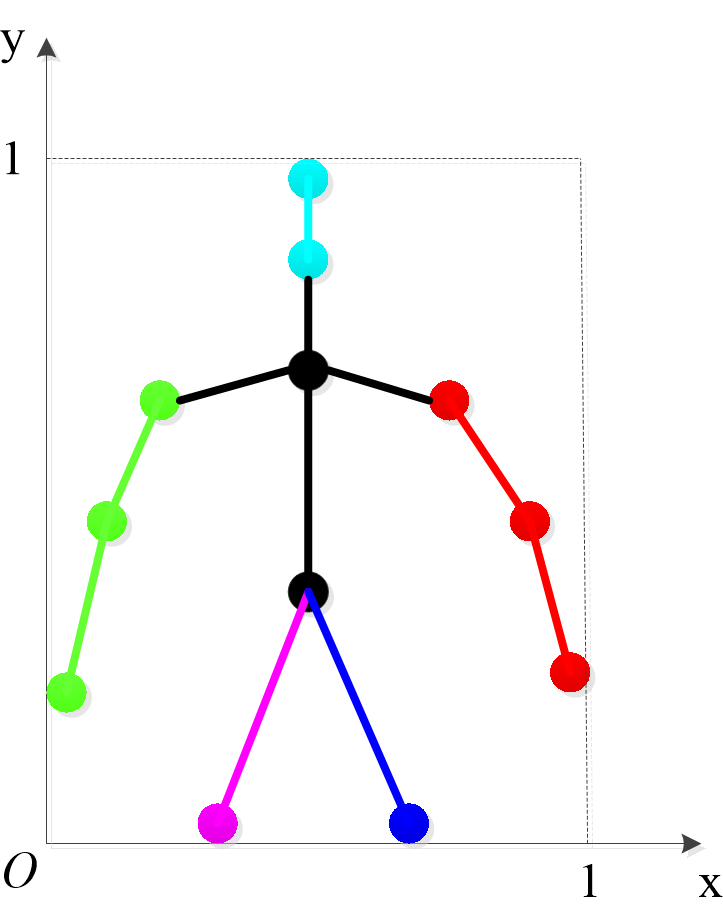}
    \end{minipage}}
  \subfigure[]{
    \centering
    \label{fig:indifail:b}
    \begin{minipage}[b]{0.15\textwidth}
      \centering
      \includegraphics[scale=0.3]{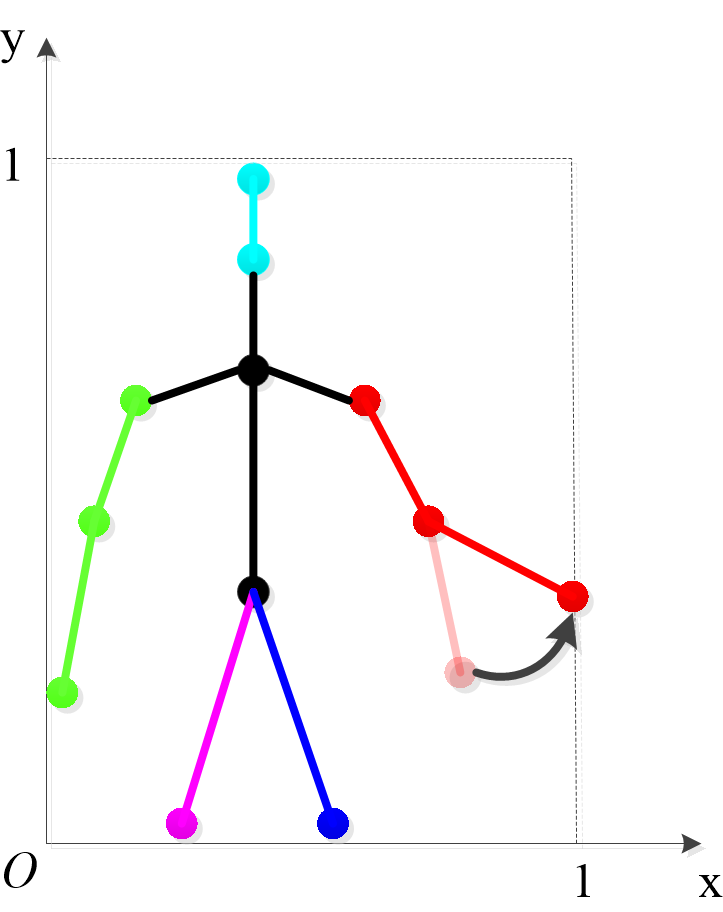}
    \end{minipage}}
  \subfigure[]{
    \centering
    \label{fig:indifail:c}
    \begin{minipage}[b]{0.15\textwidth}
      \centering
      \includegraphics[scale=0.3]{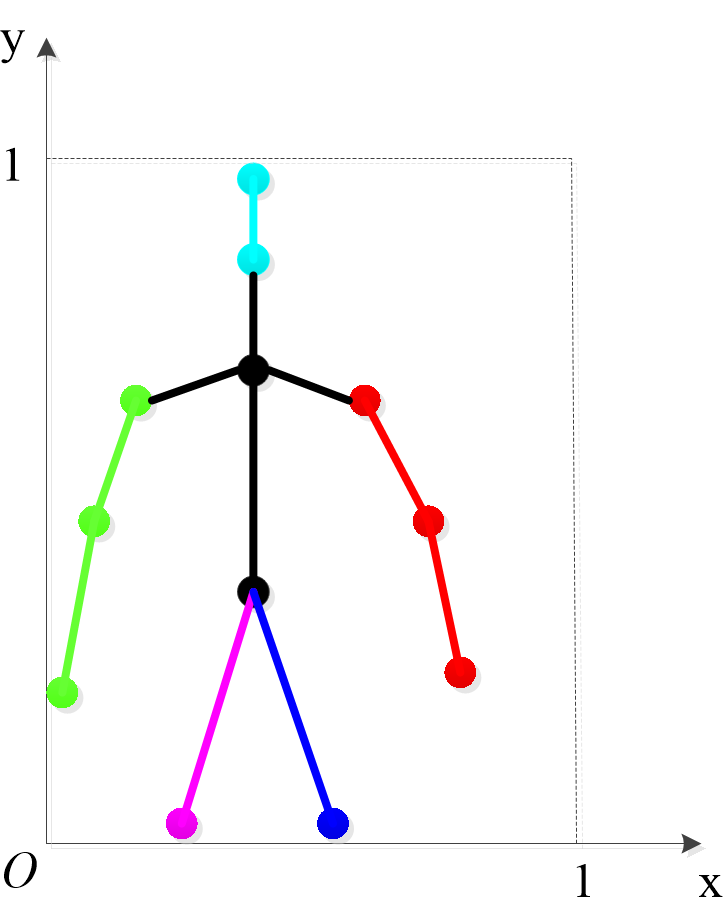}
    \end{minipage}}
  \caption{Comparison of individual normalization (a)-(b) and global normalization(b)-(c). With the same body motions illustated in (b), in individual normalization, the $x\text{-coordinates}$ of all the still keypoints rather than the moving one are changed. While global normalization correctly changes the coordinate of the moving keypoint.}
  \label{fig:indifail}
\end{figure}


Individual normalization brings all values into $[0, 1]$, which eliminates the effects of absolute body positions and scales. However, it has a severe disadvantage. For more concise explanations, the example will be raised in 2D space. As illustrated in Fig.~\ref{fig:indifail:a} and~\ref{fig:indifail:b}, assume that the keypoints have been normalized in frame $t$. In the frame $t+1$, the communicator raises his/her lower arm. We hoped that only the coordinates of `left wrist' would change in the normalized keypoints matrices. However, under the rules of individual normalization, the $x$-coordinate of `left wrist' will stay at 1, with all the other still keypoints squeezed. It's detrimental for the network to converge if the input features are not in accordance with the actual body movements.

\subsubsection{Global normalization}


To address the issue in individual normalization, we tried global normalization:
\begin{equation}\label{eq:k0tilde}
\begin{aligned}
 \widetilde{\mathbf{K}^i } = \left( {\overline {{\mathbf{K}}^i }  - {\mathbf{K}}_{\min } } \right) \odot {\mathbf{K}}_{{\rm{scale}}},
\end{aligned}
\end{equation}
where ${\mathbf{K}}_{\min }$ and ${\mathbf{K}}_{{\rm{scale}}}$ are obtained from all the keypoints matrices instead of $\overline {\mathbf{K}^i}$. As shown in Fig.~\ref{fig:indifail:b} and~\ref{fig:indifail:c}, global normalization can correctly change the coordinate of the moving keypoint.

\begin{figure} [tbp]
  \centering
  \subfigure[]{
    \centering
    \label{fig:glbfail:a} 
    \begin{minipage}[b]{0.2\textwidth}
      \centering
      \includegraphics[scale=0.3]{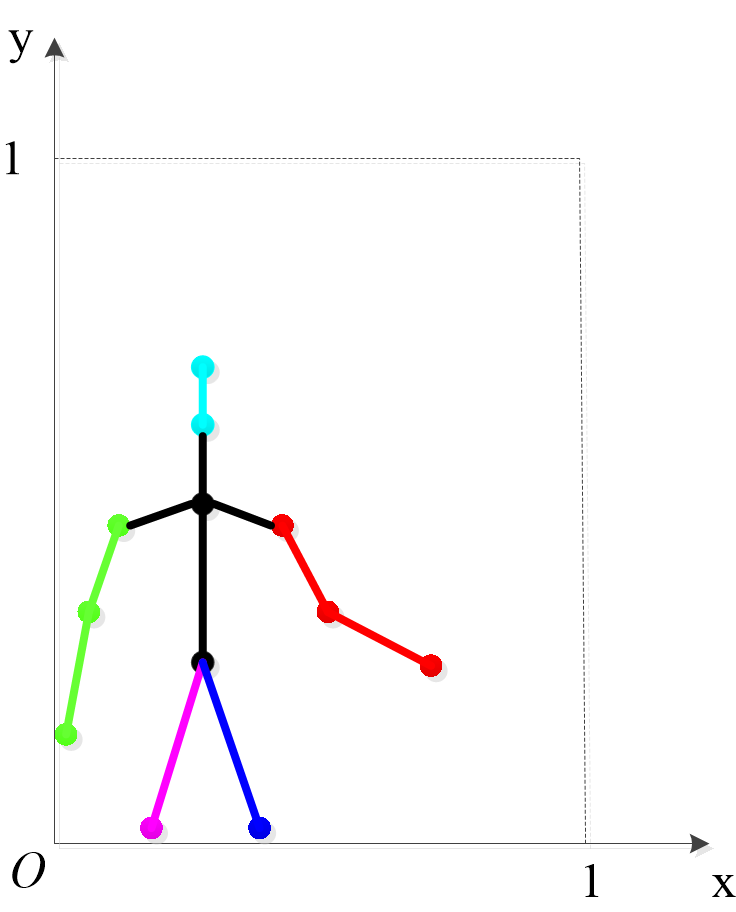}
    \end{minipage}}
  \subfigure[]{
    \centering
    \label{fig:glbfail:b}
    \begin{minipage}[b]{0.2\textwidth}
      \centering
      \includegraphics[scale=0.3]{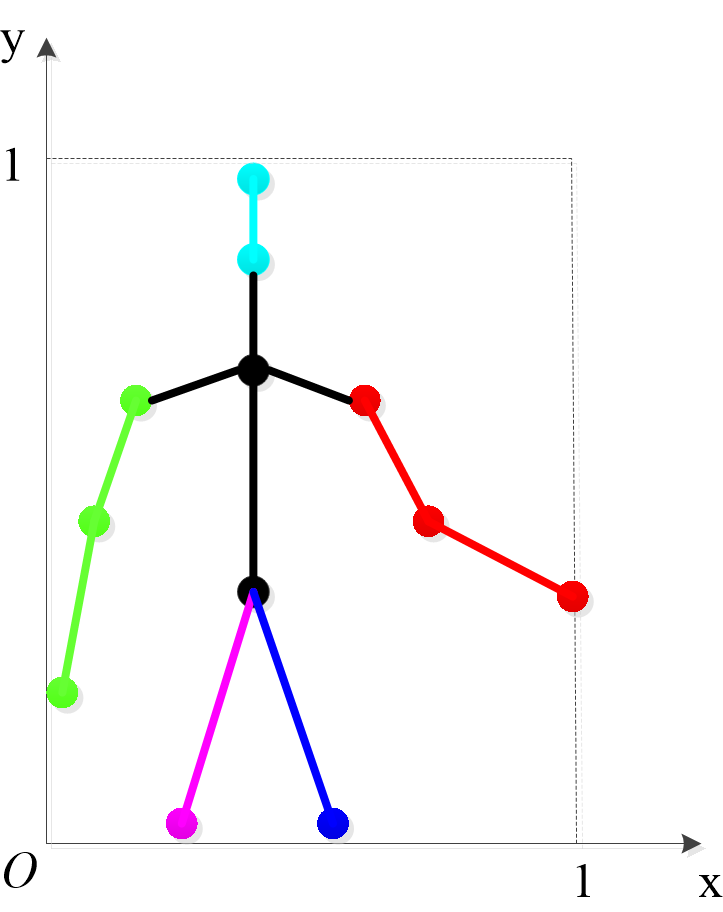}
    \end{minipage}}
  \caption{Defect of global normalization. The same body gestures in (a) and (b) have different representations due to the body scales.}
  \label{fig:glbfail}
\end{figure}

However, the global normalization cannot standardize the absolute scale, which means that the normalized values are proportional to the size of body skeleton. Fig.~\ref{fig:glbfail} shows an example that may happen in global normalization: Although two humans are acting the same body gesture, the normalized coordinates are unequal because of their body scales.

\subsubsection{Vector normalization}

\begin{figure} [tbp]
  \centering
  \subfigure[]{
    \centering
    \label{fig:vecnorm:a} 
    \begin{minipage}[b]{0.2\textwidth}
      \centering
      \includegraphics[scale=0.3]{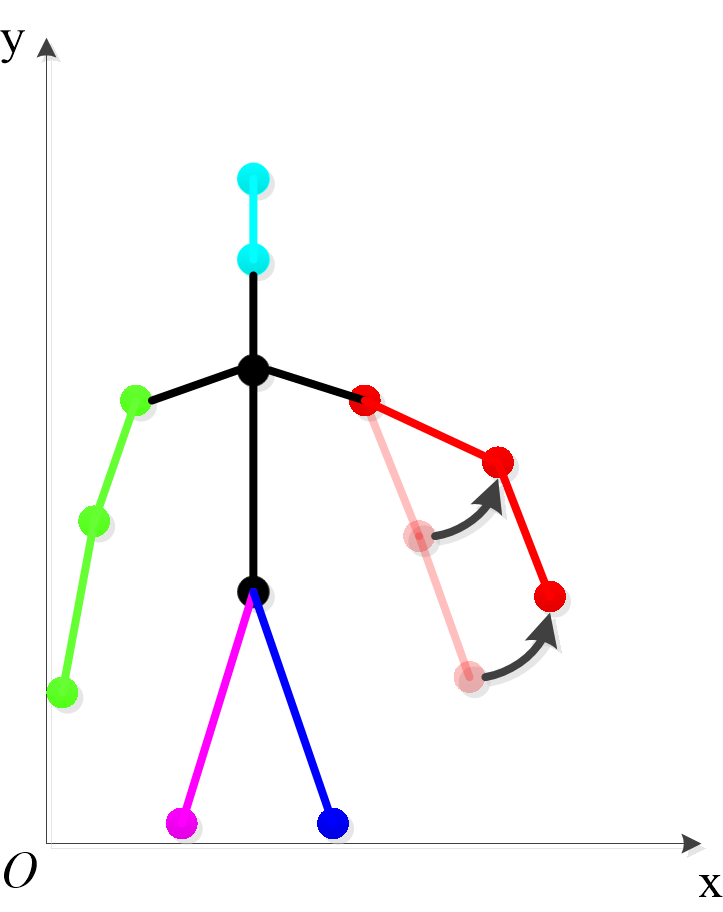}
    \end{minipage}}
  \subfigure[]{
    \centering
    \label{fig:vecnorm:b}
    \begin{minipage}[b]{0.2\textwidth}
      \centering
      \includegraphics[scale=0.3]{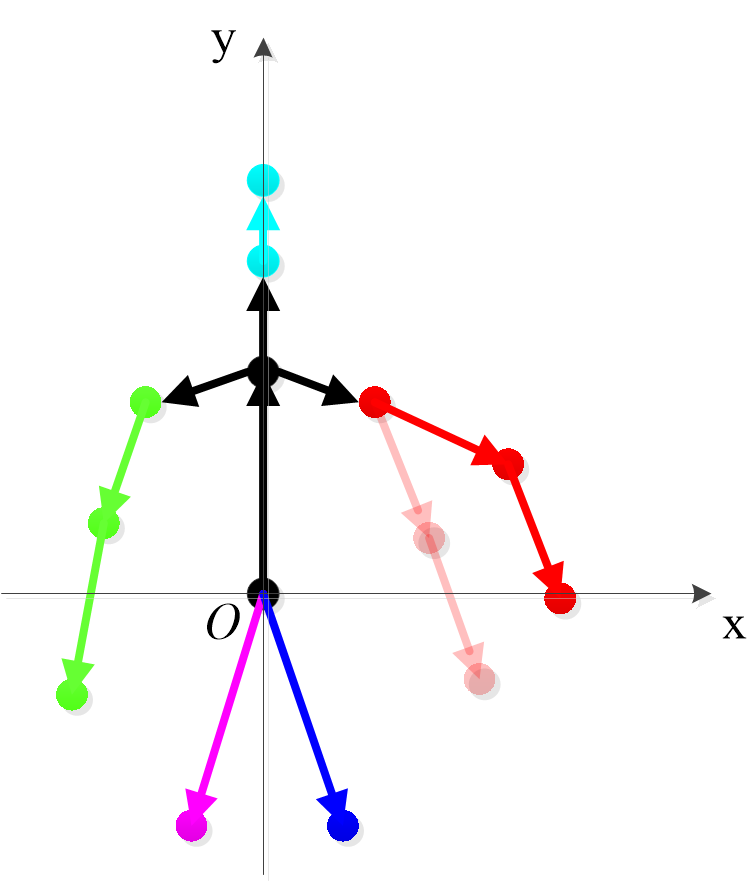}
    \end{minipage}}
  \caption{Motivation of vector normalization. (a) presents a defect of coordinate representation. (b) is a schematic diagram for vector normalization.}
  \label{fig:vecnorm}
\end{figure}

The aforementioned two normalization methods are based on the coordinate representation. In our experiments, we found coordinate representation cannot reflect the essential body movements. As the example raised in Fig.~\ref{fig:vecnorm:a}, when a communicator raises his upper arm, the coordinates of both elbow and wrist will change. However, the wrist's movement is a ripple effect of the elbow's. In other words, to make the robot imitate this body gesture, we just need to move its elbow. Therefore, the normalization method should emphasize the active keypoints movement and ignore the passive position change.

For this purpose, we exploited the vector normalization method. First, we transformed coordinates to vector representation. As illustrated in Fig.~\ref{fig:vecnorm:b}, we set belly as the center of body, whose representation is $\left( 0,0,0 \right)$. Chest and hips are connected to belly, neck and shoulders are connected to chest, head top is connected to neck, elbows are connected to the same-side shoulders, and wrists are connected to the same-side elbows. As a result, each keypoint is represented by the vector from an adjacent keypoint to itself instead of its coordinate. At last, we scaled all the vectors to unit length.

Vector normalization eliminates the effects of absolute body positions and scales. After standardizing the length of each vector, only the direction information remains. While reconstructing body gestures for the avatar, we defined a sensible length for each connection of adjacent keypoints. However, it's unnecessary for a real-world robot because the lengths of its limbs and trunk are already definite.

\section{Seq2Seq-based Listening and Speaking Models}\label{sec:model}

We adapted the sequence-to-sequence architecture~\cite{SV14} for listening and speaking phases separately. In listening phase, the input contains both the speaker's keypoint sequence and the sentence he/she said. While in speaking phase, the input contains the response sentence only. The output of both phases is a keypoint sequence, which is regarded as nonverbal feedback in the listening phase and accompaniment of utterance response in the speaking phase.

\begin{figure*}[htbp]%
\centering
\includegraphics[scale=0.5]{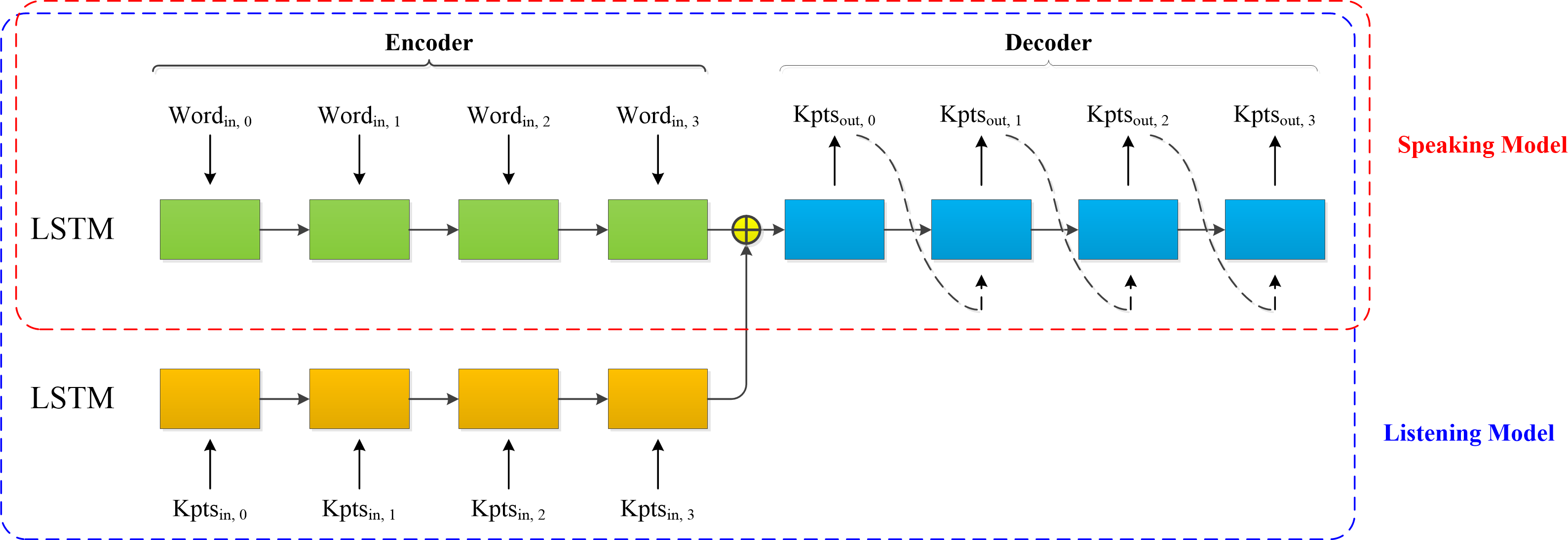}
\caption{The architecture of proposed listening and speaking models. Text encoder (colored green) is used to model a sentence. Keypoints encoder (colored orange), only exists in the listening model, is used to analyze the body gestures. Keypoints decoder (colored blue) is used to decode the latent vector and generate output keypoints sequence. Both encoder and decoder adopt LSTM layer.}
\label{fig:model}
\end{figure*}

The architecture of two models is illustrated in Fig.~\ref{fig:model}. During listening phase, two LSTMs are used to encode speaker's verbal signals and body gestures separately. Then, feature vectors from these two LSTMs are fused and decoded to generate body gestures feedback. However, during speaking phase, only the response sentence is encoded into a latent vector, which is then decoded to synthesize body gestures accompaniment.

To parse the verbal features, we implemented word embedding, which is a collection of techniques mapping words or phrases into numerical vectors. Landmark models include Word2Vec~\cite{MS13, MC13}, GloVe~\cite{PS14}, ELMo~\cite{PM18} and BERT~\cite{DC18}. Word2Vec is a computationally-efficient word embedding algorithm, which is built on either the Continuous Bag-of-Words (CBOW) model or the Skip-Gram model. GloVe introduces global matrix factorization and local context window methods to improve the performance of word embedding. ELMo and BERT both focus on the fact that the same word might have different meanings based on the context. However, ELMo adopts stack LSTM while BERT, the state-of-the-art word embedding model, uses Transformer proposed in~\cite{VA17}. In our experiments, we finetuned a pretrained Word2Vec model based on CBOW to embed each word to a vector.

The nonverbal features were also unified to vector representations. We simply flattened each keypoint matrix $\widetilde{\mathbf{K}^i }$ into a 36-D vector, and then concatenated every 10 contiguous keypoint vectors into a single one. The reason for gathering every 10 frames together is that we analyzed all the videos in the dataset and found that the average speech rate is 0.4 seconds per word. Since the videos were captured at 25 fps, body movements happening in 10 frames approximately correspond to speaking a word. Consequently, the body gestures would be represented by a sequence of 360-D vectors. To equate the dimensionality of feature vectors for both body gestures and sentences, our Word2Vec model embedded each word into a 360-D vector, thus sentences would also be represented by a sequence of 360-D vectors.

Recall the mechanism of LSTM network proposed in~\cite{HS97}. Assume the input sequence is $\left( {{\mathbf{x}}_{1}},{{\mathbf{x}}_{2}},...,{{\mathbf{x}}_{m}} \right)$, the $t\text{-th}$ LSTM unit updates the states based on the states at $t-1$:

\begin{eqnarray}\label{eq:state}
\begin{aligned}
& {{\mathbf{i}}_{t}} = \sigma \left( {{\mathbf{W}}_{i}}\left[ {{\mathbf{h}}_{t-1}},{{\mathbf{x}}_{t}} \right]+{{\mathbf{b}}_{i}} \right) \\
& {{\mathbf{f}}_{t}} = \sigma \left( {{\mathbf{W}}_{f}}\left[ {{\mathbf{h}}_{t-1}},{{\mathbf{x}}_{t}} \right]+{{\mathbf{b}}_{f}} \right) \\
& {{\mathbf{o}}_{t}} = \sigma \left( {{\mathbf{W}}_{o}}\left[ {{\mathbf{h}}_{t-1}},{{\mathbf{x}}_{t}} \right]+{{\mathbf{b}}_{o}} \right) \\
& \widetilde{{{\mathbf{c}}_{t}}} = \tanh \left( {{\mathbf{W}}_{c}}\left[ {{\mathbf{h}}_{t-1}},{{\mathbf{x}}_{t}} \right]+{{\mathbf{b}}_{c}} \right) \\
& {{\mathbf{c}}_{t}} = {{\mathbf{f}}_{t}}\odot {{\mathbf{c}}_{t-1}}+{{\mathbf{i}}_{t}}\odot \widetilde{{{\mathbf{c}}_{t}}} \\
& {{\mathbf{h}}_{t}} = {{\mathbf{o}}_{t}}\odot \tanh \left( {{\mathbf{c}}_{t}} \right),
\end{aligned}
\end{eqnarray}
where $\sigma$ denotes the sigmoid function, $\tanh$ is the hyperbolic tangent function, $\odot$ denotes the element-wise multiplication, ${{\mathbf{i}}_{t}}$, ${{\mathbf{f}}_{t}}$, ${{\mathbf{o}}_{t}}$ represent input gate, forget gate, output gate of $t$-th LSTM unit respectively. ${{\mathbf{c}}_{t}}$ and ${{\mathbf{h}}_{t}}$ are the $t$-th cell state and hidden state. $\mathbf{W}$ and $\mathbf{b}$ are trainable weights.

Therefore, two hidden vectors will be produced from the encoder during listening phase. Then they are fused by element-wise addition and passed into the decoder. During speaking phase, the encoder only produces one hidden vector, which is directly input into the decoder.

In the decoder stage, keypoint sequence is generated one at a step. The first LSTM unit receives the hidden state ${{\mathbf{h}}_{n}}$  and outputs the first prediction. Latter LSTM units take the previous prediction as input and calculate the current keypoints prediction. After the output sequence is produced, the mean squared error (MSE) will be calculated by:

\begin{equation}\label{eq:mse}
\begin{aligned}
l=\frac{1}{n}\sum\limits_{i=1}^{n}{{{\left\| \mathbf{y}_{gt}^{(i)}-\mathbf{y}_{pred}^{(i)} \right\|}_{2}}} ,
\end{aligned}
\end{equation}
where $\mathbf{y}_{gt}^{(i)}$ and $\mathbf{y}_{pred}^{(i)}$ denote the $i\text{-th}$ keypoints vector in the ground truth and predicted sequence respectively. $||\cdot |{{|}_{2}}$ denotes the Euclidean norm and $n$ is the number of LSTM units in the decoder.

\section{Experiments}\label{sec:exp}

\subsection{Implementation Details}

We implemented our seq2seq-based listening and speaking models using TFLearn~\cite{TY16}, which is a Python third-party module built on TensorFlow~\cite{TSFL}. In the training stage, network weights were optimized by minimizing the loss~\eqref{eq:mse} using Adam~\cite{ADAM}. Hyper-parameters were tuned on the validation set and the following values were applied at last: the batch size was 64, the initial learning rate was 0.01 and the training process stopped after 40000 epochs. Moreover, the LSTM layer, in both encoder and decoder, contains 7 units, which means that our models accept a maximum of 7 words for text input and 70 contiguous frames for keypoints input (recall that we packed every 10 frames in one vector). For sequences with shorter length, zero-paddings will be concatenated to their tails.

\begin{figure} [htbp]
  \centering
  \subfigure[]{
    \centering
    \label{fig:ava:a} 
    \begin{minipage}[b]{0.1\textwidth}
      \centering
      \includegraphics[scale=0.22]{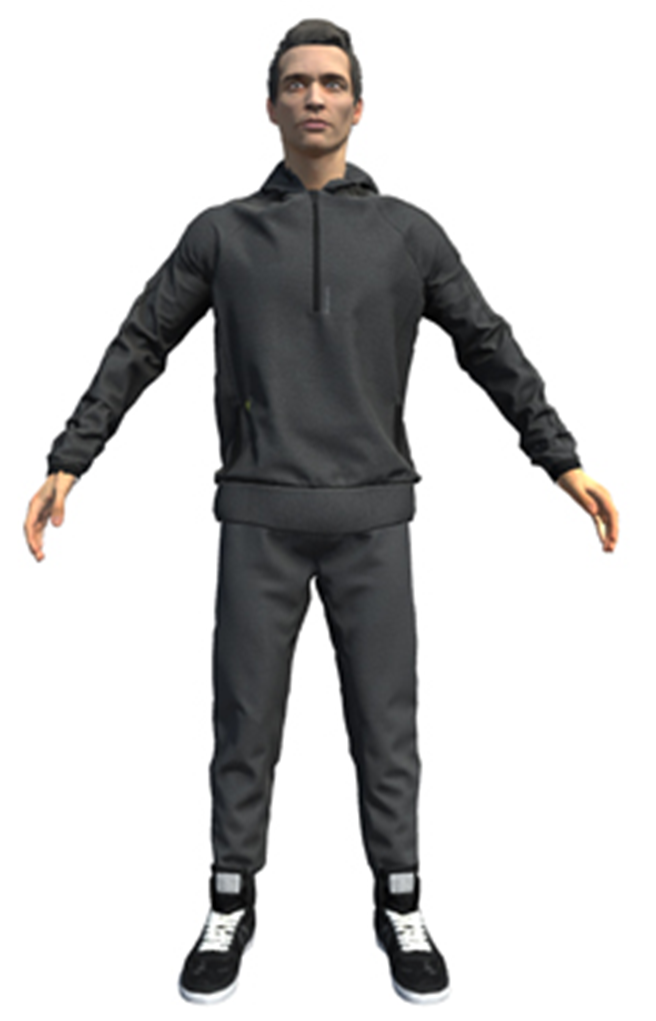}
    \end{minipage}}
  \subfigure[]{
    \centering
    \label{fig:ava:b}
    \begin{minipage}[b]{0.1\textwidth}
      \centering
      \includegraphics[scale=0.22]{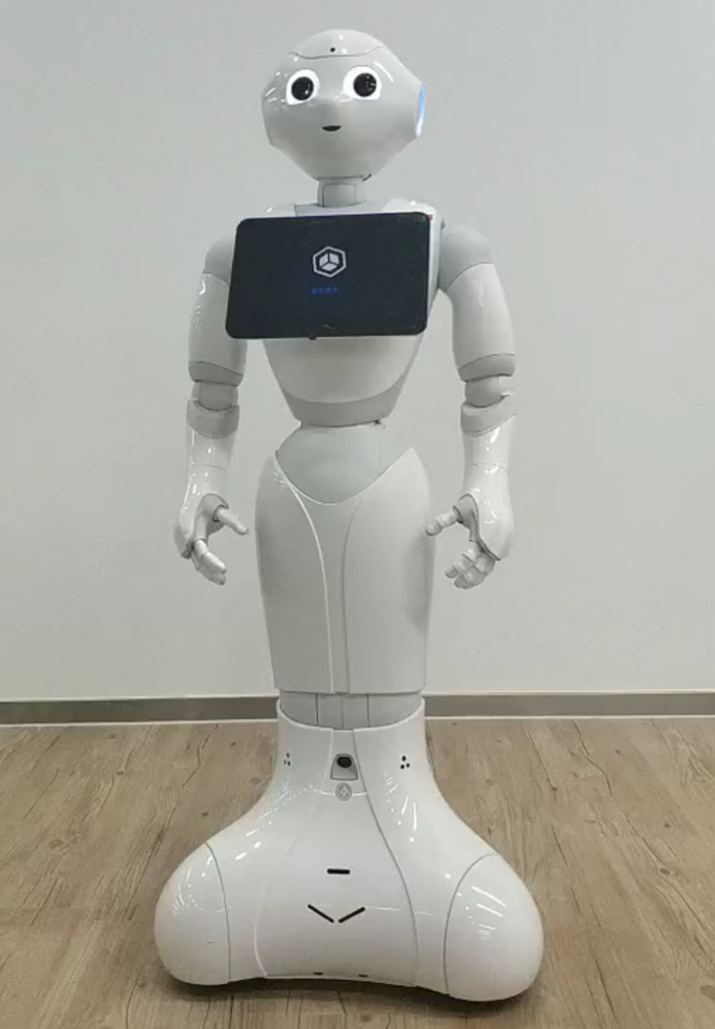}
    \end{minipage}}
  \subfigure[]{
    \centering
    \label{fig:ava:c} 
    \begin{minipage}[b]{0.1\textwidth}
      \centering
      \includegraphics[scale=0.22]{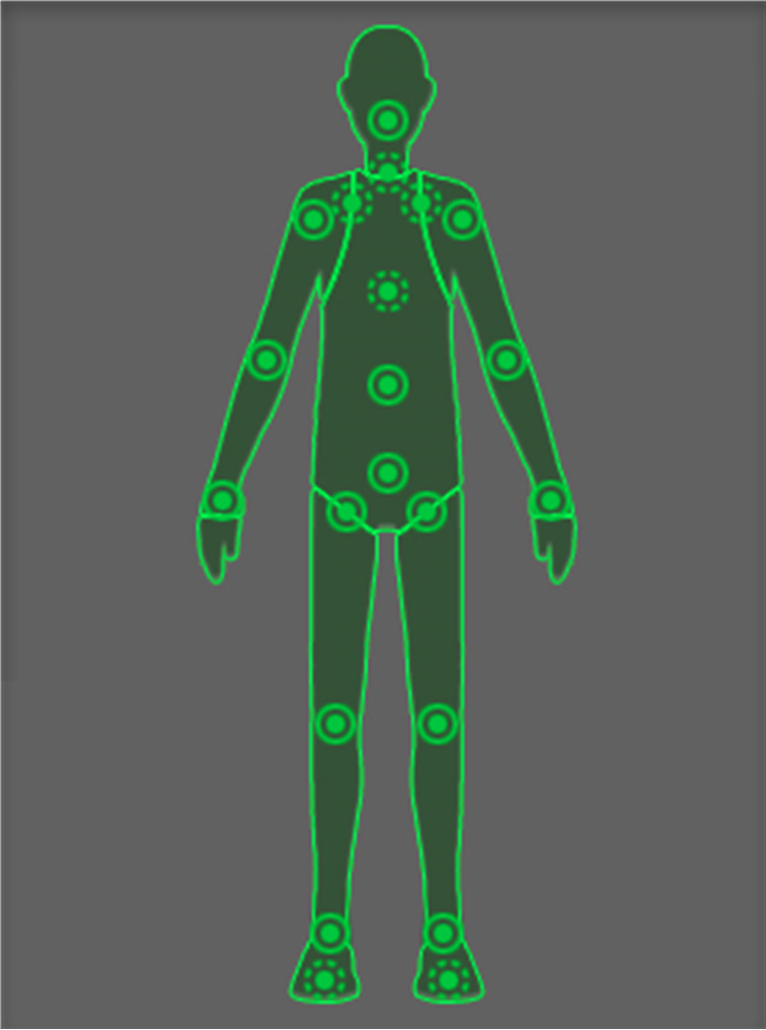}
    \end{minipage}}
  \subfigure[]{
    \centering
    \label{fig:ava:d}
    \begin{minipage}[b]{0.1\textwidth}
      \centering
      \includegraphics[scale=0.22]{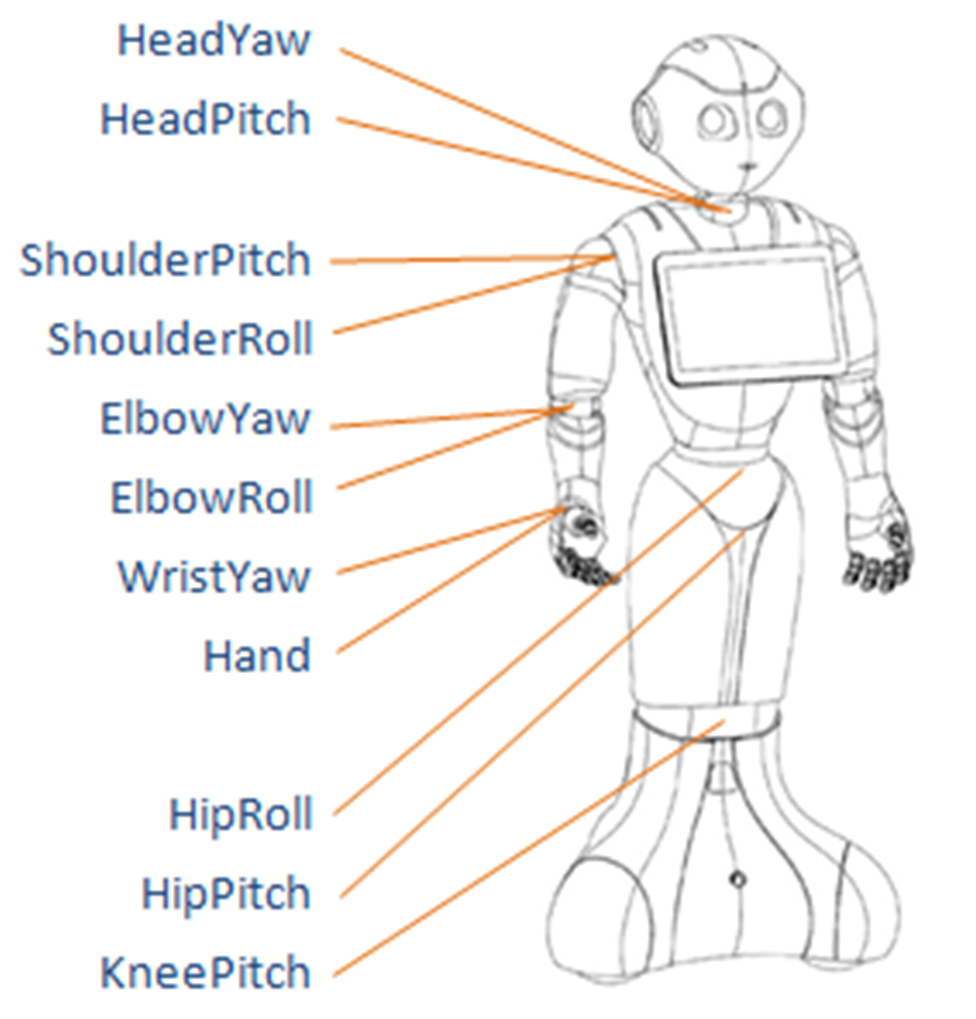}
    \end{minipage}}
  \caption{The avatar and Pepper used for body motion synthesis. (a) and (b) show the avatar and Pepper used in our experiments respectively. (c) and (d) illustrate their controllable keypoints}
  \label{fig:ava}
\end{figure}


For body motion synthesis, we reconstructed the body gestures to the avatar (Fig.~\ref{fig:ava:a}) and the humanoid robot Pepper (Fig.~\ref{fig:ava:b}) with synthesized keypoint sequence. Fig.~\ref{fig:ava:c} and \ref{fig:ava:d} show the controllable joints of avatar and Pepper~\cite{PEPPER}. First, we selected one frame every ten frames from the keypoint sequence to form a new sequence for interpolation. After that, we chose belly as the benchmark and adjusted other vectors in the new sequence to align approximately with the avatar by geometric transformation. Then we computed the angle between adjacent vectors in the first frame. After rotating all the keypoints by the corresponding angles, the avatar and Pepper would be in the pose as the first frame. Finally, we calculated the rotation angles of all the adjacent vectors between consecutive frames, interpolated values and rotated each vector so that the intelligent agents would present target body gestures smoothly based on the keypoint sequence.

\subsection{Metrics \& Evaluations}

\begin{figure*}[htbp]%
\centering
\includegraphics[scale=0.25]{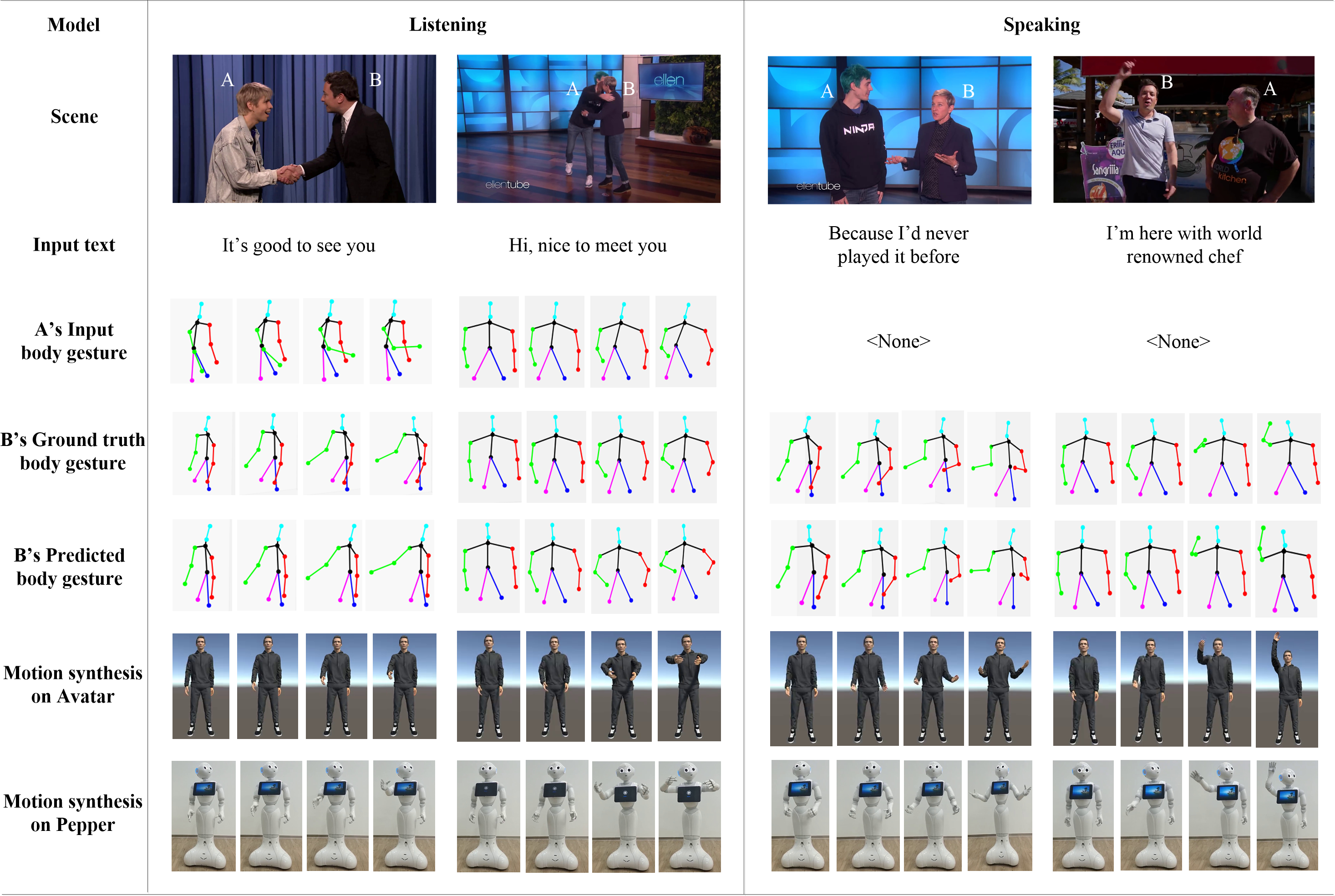}
\caption{Experimental results of our system. The listening model synthesizes appropriate body gestures based on both verbal and nonverbal cues of speaker. And the speaking model synthesizes realistic body gestures accompaniment for utterance response.}
\label{fig:result}
\end{figure*}

We applied MSE and cosine similarity to measure the accuracy of listening model and speaking model separately on the test set. For listening model, each test sample is composed of a word sequence and a keypoint sequence of the speaker as input, and a keypoint sequence of the listener as ground truth. For speaking model, the input is a word sequence of the speaker, and the ground truth is his/her keypoint sequence while speaking.

Specifically, suppose that there are $m$ test samples for the listening model, the speaker's word and keypoint sequences of each sample are input into our listening model to predict the listener's keypoint sequence. Then we use \eqref{eq:mse} to calculate loss ${{l}^{i}}$ for $i$-th sample, and the total loss $L$ among the test set is calculated by:

\begin{equation}\label{eq:loss}
\begin{aligned}
L=\frac{1}{m}\sum\limits_{i=1}^{m}{{{l}^{i}}}.
\end{aligned}
\end{equation}
And the cosine similarity ${{S}_{C}}$ is obtained as follow:

\begin{equation}\label{eq:cos}
\begin{aligned}
	{{S}_{C}}=\frac{1}{m} \sum\limits_{i=1}^{m}{\frac{\mathbf{y}_{gt}^{(i)}\cdot \mathbf{y}_{pred}^{(i)}}{\left\| \mathbf{y}_{gt}^{(i)} \right\|\left\| \mathbf{y}_{pred}^{(i)} \right\|}}.
\end{aligned}
\end{equation}
Similar evaluations are applied to the speaking model, with only the input and ground truth sequence different.

\begin{table}[tbp]
\center
  \caption{Comparison of Three Normalization Methods Based on the Loss $L$}
  \label{tb:norml}
\begin{center}
\begin{tabular}{c|c|ccc}
\hline
\multicolumn{1}{c|}{\multirow{2}{*}{Dataset}} &
\multicolumn{1}{c|}{\multirow{2}{*}{Model}} &
\multicolumn{3}{c}{Normalization Method}\\
\cline{3-5}
& & individual & global & vector\\
\hline \multirow{2}{*}{Ellen Show}&listening&0.506&0.314&\textbf{0.308}\\
&speaking&0.571&0.353&\textbf{0.336}\\
\hline \multirow{2}{*}{Tonight Show}&listening&0.525&0.324&\textbf{0.316}\\
&speaking&0.566&0.350&\textbf{0.335}\\
\hline
\end{tabular}
\end{center}
\end{table}

\begin{table}[tbp]
\center
  \caption{Comparison of Three Normalization Methods Based on the Cosine Similarity $S_c$}
  \label{tb:normsc}
\begin{center}
\begin{tabular}{c|c|ccc}
\hline
\multicolumn{1}{c|}{\multirow{2}{*}{Dataset}} &
\multicolumn{1}{c|}{\multirow{2}{*}{Model}} &
\multicolumn{3}{c}{Normalization Method}\\
\cline{3-5}
& & individual & global & vector\\
\hline \multirow{2}{*}{Ellen Show}&listening&0.983&0.988&\textbf{0.992}\\
&speaking&0.976&0.978&\textbf{0.984}\\
\hline \multirow{2}{*}{Tonight Show}&listening&0.979&0.985&\textbf{0.989}\\
&speaking&0.975&0.980&\textbf{0.985}\\
\hline
\end{tabular}
\end{center}
\end{table}

We compared three normalization methods on both datasets. It can be seen from Table~\ref{tb:norml} and Table~\ref{tb:normsc} that vector normalization method outperforms individual and global normalization methods on both MSE and cosine similarity metrics in both datasets. Therefore, we finally adopted vector normalization method in our experiments.

\begin{table}[tbp]
\center
  \caption{Comparison of Fixed and Flexible Learning Targets}
  \label{tb:lt}
\begin{center}
\begin{tabular}{c|c|cccc}
\hline
\multicolumn{1}{c|}{\multirow{2}{*}{Metrics}} &
\multicolumn{1}{c|}{\multirow{2}{*}{Model}} &
\multicolumn{4}{c}{Learning Target}\\
\cline{3-6}
& & Ellen & Jimmy & both & guests\\
\hline \multirow{2}{*}{$L$}&listening&\textbf{0.308}&0.316&0.347&0.379\\
&speaking&0.336&\textbf{0.335}&0.385&0.423\\
\hline \multirow{2}{*}{$S_c$}&listening&\textbf{0.992}&0.989&0.985&0.981\\
&speaking&0.984&\textbf{0.985}&0.978&0.970\\
\hline
\end{tabular}
\end{center}
\end{table}

\begin{table}[htbp]
\center
  \caption{Evaluation on Separate Keypoint Vector}
  \label{tb:kv}
\begin{center}
\begin{tabular}{l|cc|cc}
\hline
\multicolumn{1}{c|}{\multirow{2}{*}{Keypoint Vector}} &
\multicolumn{2}{c|}{Listening} &
\multicolumn{2}{c}{Speaking}\\
\cline{2-5}
& $L$ & $S_c$ & $L$ & $S_c$ \\
\hline
\qquad\;\;\; belly (the center) & 0.000 & 1.000 & 0.000 & 1.000 \\
\hline
\qquad\quad\,\, belly $\rightarrow$ chest & 0.085 & 0.997 & 0.092 & 0.991 \\
\hline
\qquad\quad\; chest $\rightarrow$ neck & 0.082 & 0.999 & 0.092 & 0.995 \\
\hline
\qquad\quad\;\, neck $\rightarrow$ head top & 0.087 & 0.997 & 0.093 & 0.990 \\
\hline
\qquad\quad\; chest $\rightarrow$ left shoulder & 0.090 & 0.992 & 0.100 & 0.986 \\
\hline
\, left shoulder $\rightarrow$ left elbow & 0.096 & 0.986 & 0.104 & 0.979 \\
\hline
\quad\: left elbow $\rightarrow$ left wrist & 0.107 & 0.983 & \textbf{0.118} & 0.972 \\
\hline
\qquad\quad\; chest $\rightarrow$ right shoulder & 0.092 & 0.989 & 0.099 & 0.983 \\
\hline
right shoulder $\rightarrow$ right elbow & 0.096 & 0.980 & 0.107 & 0.970 \\
\hline
\quad right elbow $\rightarrow$ right wrist & \textbf{0.110} & \textbf{0.972} & 0.116 & \textbf{0.963} \\
\hline
\qquad\quad\,\, belly $\rightarrow$ left hip & 0.085 & 0.997 & 0.092 & 0.992 \\
\hline
\qquad\quad\,\, belly $\rightarrow$ right hip & 0.087 & 0.998 & 0.096 & 0.991 \\
\hline
\end{tabular}
\end{center}
\end{table}

We designed an experiment to exchange the input and ground truth sequence, thus the model attempted to learn the body gestures of the guest instead of the host. As shown in Table~\ref{tb:lt}, it is proved that the persona had huge effects on our models. In Table~\ref{tb:kv}, we compared the accuracy of each keypoint vector separately. It is shown that the vector connecting elbow and wrist has the largest deviation, which might be caused by the frequent movements of the lower arm when a human is talking.

In Fig.~\ref{fig:result}, the experimental results of four examples are illustrated to show the keypoints comparison between ground truth and our predictions, as well as the synthesized body gestures on both the avatar and the Pepper.

\section{Conclusion}\label{sec:conc}

In this paper, we propose a novel body gesture interaction system for more realistic human-robot conversations. The seq2seq architecture is adapted to a listening model and a speaking model for body gesture synthesis in corresponding conversational phases. Both models utilize LSTM network to encode and decode the text and body gestures represented by 12 crucial upper-body keypoints, which are extracted, 3D-transformed, rotated and normalized by the gesture parsing module. Our models are trained by substantial talk show videos downloaded from Youtube and evaluated by the metrices of MSE and cosine similarity. Synthesized body gestures are reconstructed to the avatar and Pepper using the keypoints vector sequence predicted by the models. Experimental results show that the proposed models are able to learn human's body gestures during both listening and speaking phases in conversation, and the proposed interaction system is possible to provide natural human-robot conversation experiences. In future, real-time conversation model based on the proposed system will be investigated.










\bibliographystyle{IEEEtran}
\bibliography{IEEEabrv,IEEEexample}

\end{document}